\documentclass{article}

\usepackage{microtype}
\usepackage{graphicx}
\usepackage{subcaption}
\usepackage{booktabs}
\usepackage{multirow}
\usepackage{array}
\usepackage{xparse}

\usepackage[preprint]{style_sheets/icml2026}

\usepackage{amsmath}
\usepackage{amssymb}
\usepackage{mathtools}
\usepackage{amsthm}
\usepackage{arydshln}
\usepackage[table]{xcolor}
\usepackage{xfp} 

\usepackage{seqsplit}
\usepackage{hyperref}

\usepackage{xurl} 

\usepackage{hyperref}
\usepackage[capitalize,noabbrev]{cleveref}

\theoremstyle{plain}

\theoremstyle{definition}

\theoremstyle{remark}

\usepackage[textsize=tiny]{todonotes}

\usepackage[utf8]{inputenc} 
\usepackage[T1]{fontenc}    
\usepackage{amsfonts}       
\usepackage{nicefrac}       
\usepackage[normalem]{ulem}
\usepackage{makecell} 
\usepackage{natbib} 
\usepackage{pifont} 
\usepackage{tcolorbox}
\usepackage{tikz}

\usepackage{listings}
\usepackage{seqsplit}

\usepackage{xcolor}
\definecolor{darkred}{RGB}{139,0,0}

\usepackage{tcolorbox}
\tcbuselibrary{breakable}
\tcbuselibrary{skins}
\usepackage{cuted}
\usepackage[frozencache,cachedir=minted-cache]{minted}
\lstdefinestyle{promptstyle}{
  basicstyle=\ttfamily\footnotesize,
  columns=fullflexible,
  breaklines=true,
  breakatwhitespace=true,
  keepspaces=true,
  showstringspaces=false,
  frame=single,
  rulecolor=\color{black!20},
}
\usepackage{caption}
\newcommand{\thickuline}[2][0.8pt]{%
  \begingroup
  \setbox0=\hbox{#2}%
  \leavevmode
  \vbox{%
    \box0
    \kern1pt
    \hrule height #1
  }%
  \endgroup
}

\newcommand{\DATANAME}{\textsc{TaoBench}}
\newcommand{\DATANAMEMathlib}{\textsc{TaoBenchMathLib}}
\selectcolormodel{rgb}
\definecolor{heatred}{RGB}{248,195,195}    
\definecolor{heatgreen}{RGB}{190,235,200}  

\newcommand{\HeatCellMM}[3]{%
  \begingroup
  \def\HCval{#1}\def\HCmin{#2}\def\HCmax{#3}%
  \def\HCmissing{--}%
  \ifx\HCval\HCmissing
    --%
  \else
    \edef\HCden{\fpeval{\HCmax-\HCmin}}%
    \edef\HCabsden{\fpeval{abs(\HCden)}}%
    \ifdim \HCabsden pt < 0.000000001pt
      #1%
    \else
      \edef\HCt{\fpeval{min(1,max(0,(\HCval-\HCmin)/(\HCden)))}}%
      \edef\HCd{\fpeval{abs(\HCt-0.5)/0.5}}%
      \edef\HCs{\fpeval{round(min(100,max(0,100*(\HCd^3.6))))}}%
      \edef\HCs{\fpeval{ifthenelse(\HCs<6,0,\HCs)}}%
      \ifdim \HCt pt < 0.5pt
        \cellcolor{heatred!\HCs!white}#1%
      \else
        \cellcolor{heatgreen!\HCs!white}#1%
      \fi
    \fi
  \fi
  \endgroup
}

\icmltitlerunning{\DATANAME: Do Automated Theorem Prover LLMs Generalize Beyond MathLib?}

\begin{document}

\twocolumn[

  \icmltitle{\DATANAME: Do Automated Theorem Prover LLMs Generalize Beyond MathLib?}

  \vspace{-3.1mm}



  \icmlsetsymbol{equal}{*}
  \icmlsetsymbol{senior}{$\dagger$}

  \begin{icmlauthorlist}
    \icmlauthor{Alexander K Taylor}{equal,ucla}
    \icmlauthor{Junyi Zhang}{equal,ucla}
    \icmlauthor{Ethan Ji}{ucla}
    \icmlauthor{Vigyan Sahai}{ucsb}
    \icmlauthor{Haikang Deng}{ucla}
    \icmlauthor{Yuanzhou Chen}{ucla}
    \icmlauthor{Yifan Yuan}{ucla}
    \icmlauthor{Di Wu}{ucla}
    \icmlauthor{Jia-Chen Gu}{ucla}
    \icmlauthor{Kai-Wei Chang}{senior,ucla}
    \icmlauthor{Nanyun Peng}{senior,ucla}
    \icmlauthor{Amit Sahai}{senior,ucla}
    \icmlauthor{Wei Wang}{senior,ucla}
  \end{icmlauthorlist}

  \icmlaffiliation{ucla}{University of California, Los Angeles}
  \icmlaffiliation{ucsb}{University of California, Santa Barbara}

  \icmlcorrespondingauthor{Alexander K Taylor}{ataylor2@g.ucla.edu}
  \icmlcorrespondingauthor{Junyi Zhang}{junyizhang2002@g.ucla.edu}

  \icmlkeywords{Machine Learning, ICML}

  \begin{center}
  \textbf{\textbf{Website: \textcolor{darkred}{\url{https://TaoBench.github.io}}}}
  \end{center}

  \vspace{-3mm}

  \vskip 0.3in
]

\printAffiliationsAndNotice{* Equal contribution. $\dagger$ Senior authors served as equal advisors and are listed alphabetically.}




\begin{abstract}

Automated theorem proving (ATP) benchmarks largely consist of problems formalized in MathLib, so current ATP training and evaluation are heavily biased toward MathLib's definitional framework. However, frontier mathematics is often exploratory and prototype-heavy, relying on bespoke constructions that deviate from standard libraries.
In this work, we evaluate the robustness of current ATP systems when applied to a novel definitional framework, specifically examining the performance gap between standard library problems and bespoke mathematical constructions. 
We introduce \DATANAME{}, an undergraduate-level benchmark derived from Terence Tao's \textit{Analysis I}, 
which formalizes analysis by constructing core mathematical concepts from scratch, without relying on standard Mathlib definitions, as well as by mixing from-scratch and MathLib constructions.
For fair evaluation, we build an agentic pipeline that automatically extracts a compilable, self-contained local environment for each problem. To isolate the effect of definitional frameworks, we additionally translate every problem into a mathematically equivalent Mathlib formulation, yielding paired \DATANAMEMathlib{} statements for direct comparison.
While state-of-the-art ATP models perform capably within the MathLib framework, performance drops by an average of roughly 26\% on the definitionally equivalent Tao formulation.
This indicates that the main bottleneck is limited generalization across definitional frameworks rather than task difficulty.
\DATANAME{} thus highlights a gap between benchmark performance and  applicability, and provides a concrete foundation for developing and testing provers better aligned with research mathematics.

\end{abstract}
\section{Introduction}



\vspace{2mm}

\begin{figure}[t]
  \centering
  \includegraphics[width=0.48\textwidth]{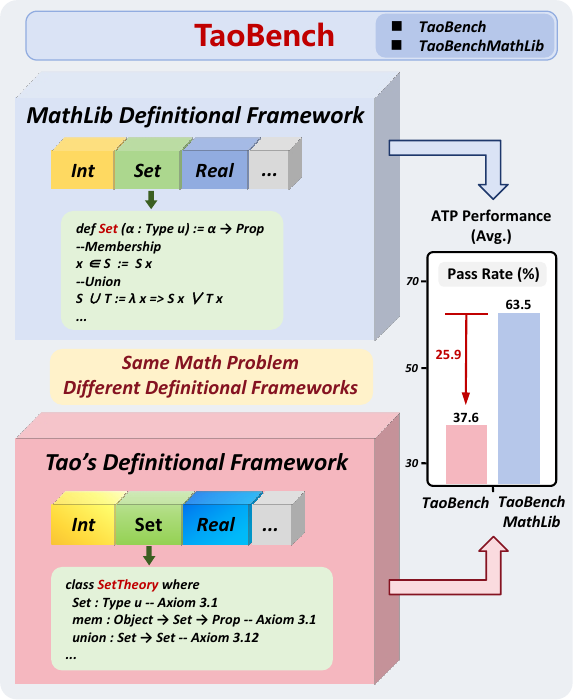}
  \caption{
  Tao's \textit{Analysis I} constructs several mathematical structures in LEAN from scratch, often using formulations that diverge significantly from their equivalents in MathLib. \DATANAME{} provides mathematically equivalent MathLib counterparts, and reveals a substantial degradation in model performance when navigating Tao’s specific definitional framework.
  }
  \label{fig:teaser_figure}
  \vspace{-4mm}
\end{figure}

\begin{figure*}[t]
  \centering
  \includegraphics[width=0.98\textwidth]{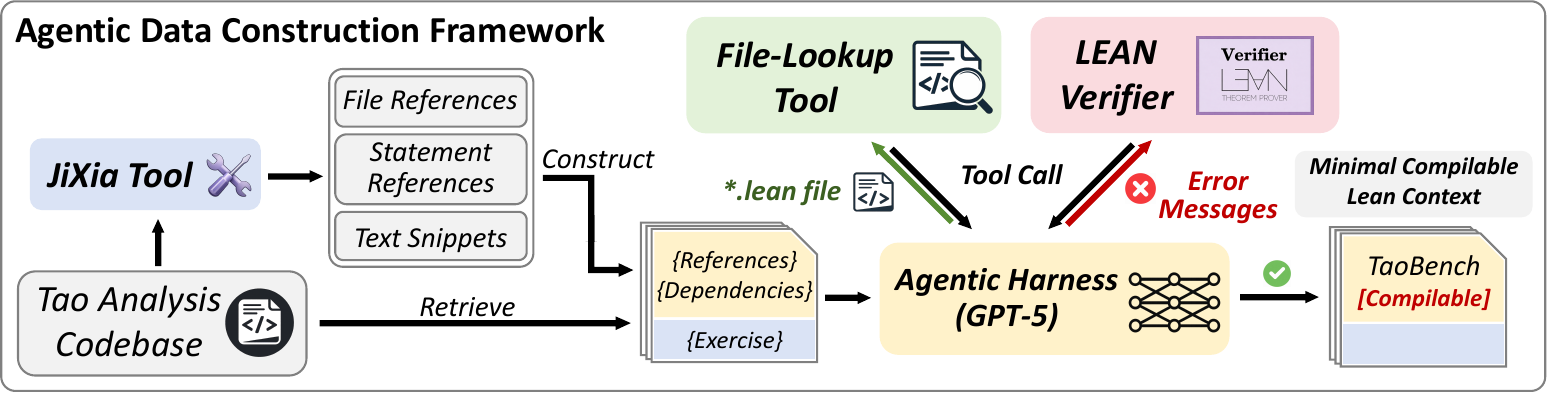}
  \caption{
  Agentic framework for automatically extracting a compilable, self-contained local environment from a formalized textbook. The construction of references and dependencies for each exercise is performed using the JiXia tool. An agentic harness, equipped with a file-lookup tool and a Lean verifier, then iteratively builds a compilable, self-contained local environment.
  }
  \label{fig:agentic_tao_version_pipeline}
  \vspace{-4mm}
\end{figure*}

Applying large language reasoning models (LLMs) to formally verifiable programming languages like Lean 4 has emerged as a critical frontier for advancing the mathematical reasoning abilities of AI.
Accordingly, several recent works have focused on training specialized automated theorem proving (ATP) LLMs to reason in Lean and have achieved high solve rates on formalized benchmarks, such as MiniF2F~\cite{goedel_prover_v2, deepseek_prover_v2, kimina_prover, miniF2F}.

Current datasets for training and evaluating ATP models~\cite{kimina_prover, deepseek_prover, deepseek_prover_v2, goedel_prover, goedel_prover_v2, BFS_prover_V2} are built almost entirely within the definitional framework of MathLib
~\cite{miniF2F, putnamBench, proofNet, AIME_dataset, herald}, 
the largest and de facto standard mathematical library for Lean 4~\cite{mathlib4}. 
However, exploratory mathematical research often ventures beyond the boundaries of MathLib and  requires problem-specific definitions and constructions to prototype concepts and guide proofs~\cite{tao_blueprint, brownian_motion_in_lean, formalizing_class_field_theory}. 
This introduces a distribution shift between the definitional framework used in public SOTA ATP model training and the frameworks to which they are applied. 
As such, current benchmarks offer limited insight into whether ATP models can generalize across such definitional frameworks.

We introduce \DATANAME{}, the first benchmark designed to measure ATP robustness beyond MathLib by targeting generalization to a novel definitional framework. 
\DATANAME{} is grounded in 150 exercises from Terence Tao's Lean formalization of Analysis I~\cite{Tao2025_AnalysisI_LeanRepo}. This formalization develops core concepts of analysis from first principles rather than relying on standard MathLib definitions, and in some cases combines from-scratch constructions with MathLib components.
To ensure fair evaluation, we develop an agentic extraction pipeline that automatically packages each problem with a self-contained, compilable local environment extracted from the formalized textbook, requiring no external imports or references. 
Moreover, we construct a paired control: for each Tao-style problem, we construct a mathematically equivalent MathLib formulation, which we title \DATANAMEMathlib{}. 
This paired representation isolates the effect of definitional frameworks from the difficulty of each problem, enabling a direct evaluation of impact on model performance.

We evaluate several state-of-the-art ATP models and find that, while these models solve analysis exercises reliably on \DATANAMEMathlib{}, performance drops by approximately 26\% on average on the same problems in \DATANAME{}. 
The observed gap implies that training in the MathLib definitional framework does not reliably transfer to unfamiliar definitional formalisms, even when the underlying mathematics is equivalent. 
We believe \DATANAME{} exposes a key hurdle towards applying public SOTA ATP models to research mathematics, and provides a concrete testbed for aligning future prover models with real-world mathematical workflows.

Our main contributions are as follows: (1) We propose \DATANAME{}, the first Lean benchmark targeting generalization beyond MathLib via a new definitional framework. 
(2) We provide \DATANAMEMathlib{}, mathematically equivalent MathLib translations of the problems in \DATANAME{} and identify a substantial generalization gap in existing ATP models.
(3) We introduce two pipelines that together automatically extract self-contained, compilable local environments for problems from a large, formalized project and provide equivalent formalizations in MathLib syntax,
which enables scalable, high-fidelity generation of mathematically grounded training data  for future models.

\section{Benchmark Construction}

\begin{figure*}[t]
  \centering
  \includegraphics[width=0.98\textwidth]{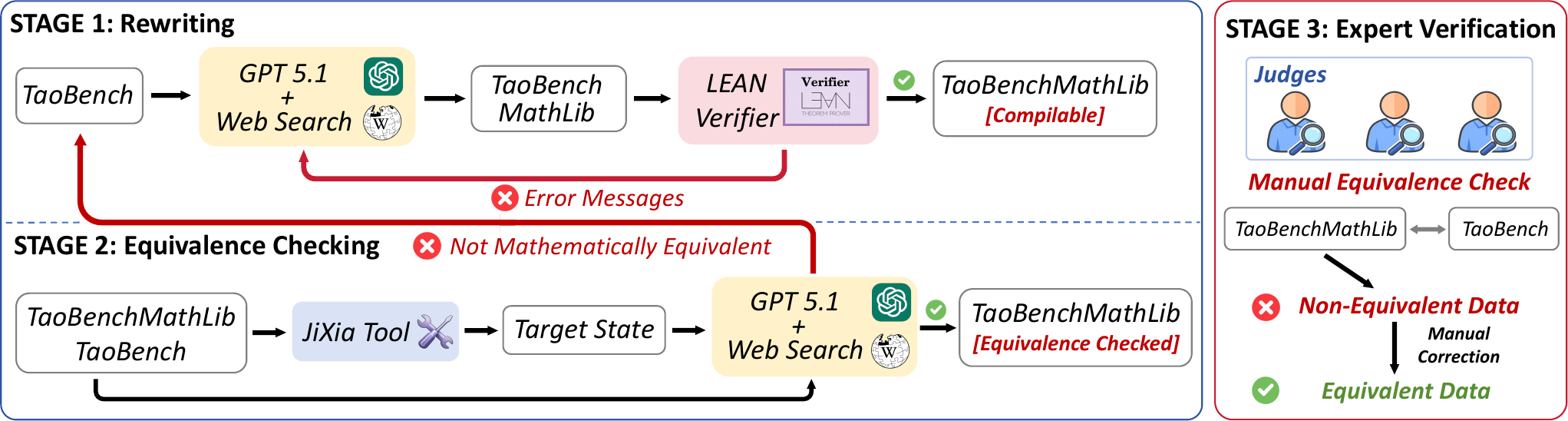}
  \caption{
   Translation pipeline for converting exercises from Tao's definitional framework to MathLib's. The rewriting and equivalence checking stages ensure the compilability and mathematical equivalence of \DATANAMEMathlib{}, followed by expert verification for additional assurance.
  }
  \label{fig:automated_translation_pipeline}
  \vspace{-4mm}
\end{figure*}



Research in mathematics often requires introducing novel axioms, definitions, and constructs that are not yet in formalized Lean libraries. Characterizing the efficacy of public ATP models in an exploratory research environment requires formalized definitional frameworks distinct from MathLib, but with enough conceptual overlap to allow models to transfer their proving capacity.
Thus, evaluating the robustness of ATP models to these novel environments requires grounding evaluation in a formal framework distinct from MathLib, while preserving enough conceptual overlap to test the transfer of existing proving capabilities.

Tao's Lean formalization of \emph{Analysis I} defines many standard mathematical objects using custom inductive types or structures for pedagogical clarity, adopts naming schemes that differ from MathLib conventions, and organizes definitions within section- and subsection-level namespaces. Additionally, some concepts are defined as fields of larger structures rather than as standalone definitions. As a result, it becomes possible that even mathematically elementary statements could become out-of-distribution to models trained primarily on MathLib-style Lean code.

To enable a fair and interpretable evaluation of model capabilities on undergraduate-level analysis, we construct twin versions of the benchmark. The first version, \DATANAME{}, evaluates models directly on formalized textbook problems and provides all relevant definitions and references necessary to correctly state the problem in a fully compilable, in-context environment. This setting is designed to assess how well models function on non-MathLib formalizations when sufficient\footnote{Sufficient in the sense that the original intent of the problem is preserved. Models often have to reference material outside the context for successful proofs.} contextual information is provided. The second version, \DATANAMEMathlib{} consists of MathLib-syntax translations of the same exercises, enabling an in-distribution evaluation that isolates mathematical difficulty from representational mismatch.

\newcommand{\TaoMin}{20.67}
\newcommand{\TaoMax}{48.67}
\newcommand{\MathMin}{53.30}
\newcommand{\MathMax}{80.00}

\begin{table*}[!t]
\centering


\caption{Model Performance on \DATANAME{} and \DATANAMEMathlib{}. Results are reported as pass rates on 150 problems. 
For the evaluation results of tree-search-based models, please see Appendix~\ref{appendix:Discussion-BFS-Prover-Evaluation} for details.}

\normalsize

\renewcommand{\arraystretch}{1}

\begin{tabular}{
p{0.36\textwidth}
>{\centering\arraybackslash}p{0.12\textwidth}
>{\centering\arraybackslash}p{0.10\textwidth}
>{\centering\arraybackslash}p{0.12\textwidth}
>{\centering\arraybackslash}p{0.10\textwidth}
}
\toprule
\multirow{2}{*}{\textbf{Model}} &
\multicolumn{2}{c}{\DATANAME{}} &
\multicolumn{2}{c}{\DATANAMEMathlib{}} \\
\cmidrule(lr){2-3} \cmidrule(lr){4-5}
& \textbf{Pass@128} & \textbf{(\%)} & \textbf{Pass@128} & \textbf{(\%)} \\
\midrule

DeepSeek-Prover-V2-7B (w/o COT)        & 57/150  & 38.00 & 82/150 & 54.67 \\
DeepSeek-Prover-V2-7B                  & \underline{62/150}  & \underline{41.33} & 104/150 & 69.33 \\
Goedel-Prover-V2-8B                    & 56/150  & 37.33 & \underline{106/150} & \underline{70.67} \\
Kimina-Prover-8B                       & 33/150  & 22.00 & 75/150  & 50.00 \\
\midrule
Goedel-Prover-V2-32B                   & \textbf{74/150}  & \textbf{49.33} & \textbf{109/150} & \textbf{72.67} \\

\bottomrule
\end{tabular}

\renewcommand{\arraystretch}{1.0}

\label{tab:tao-mathlib-results}

\vspace{-4mm}
\end{table*}


\subsection{\DATANAME{}}




Constructing a benchmark that preserves the exact textbook formalizations while producing a self-contained succinct compilable Lean snippet presents significant technical challenges for current LLMs. We find that naïvely importing entire chapters or sections of the textbook leads to excessive context length, compilation failures, and unfaithful reconstructions of the original problem and its intent. We observe that without explicit targets, modern frontier LLMs struggle to identify \textit{all} information critical for compilation over large textbook contexts containing extensive irrelevant information. 
Furthermore, querying frontier LLMs with the objective of compilation may succeed in compilation while silently changing the intended meaning. For instance, the model may introduce a local alias that collapses one or more definitions and, in so doing, trivializes the goal, e.g., \texttt{abbrev Sequence.IsCauchy := True}. This necessitates a pipeline that provides grounded reference information to the model, to avoid hallucinating and trivializing statements, as well as allowing for multiple compilation attempts. Thus, we introduce an agentic framework that retrieves the minimal context from the textbook and iteratively attempts compilation and fixes issues until it achieves successful compilation.

Individual textbook exercises typically depend on a nontrivial subset of previously defined formalized theorems, lemmas, definitions, and notation, often spread across multiple subsections. This introduces an issue for context construction for end-to-end proof generation prompts to ATP models: providing too little context may result in missing useful information, while providing too much context introduces substantial noise. We approach this by retrieving only the definitions strictly required to faithfully compile each exercise\footnote{We leave defining, identifying, and retrieving \textit{all useful} contextual information to future work.}. We accomplish this using the JiXia static Lean analysis tool~\cite{jixiaGitHub} to recursively retrieve all references of a benchmark problem.\footnote{We extend JiXia to support retrieval of \texttt{structure} declarations in addition to constants and theorems.} These references are provided in the context-construction prompt for each problem along with the file of origin to ground the context construction to a fixed set of definitions, mitigating model tendencies to hallucinate or reformulate statements.

However, retrieval based on the JiXia table alone is insufficient. The JiXia tool extracts information during the elaboration phase of compilation, and thus does not capture sugared syntax (ex. postfixes), as these constructs are reduced into their underlying definition when retrieved. Furthermore, the Lean contexts constructed after single query attempts frequently fail compilation checks due to missing information or simple syntax errors. Thus, we introduce a file-lookup tool into the agent that allows the model to retrieve source files corresponding to one or more JiXia-identified references, enabling access to the local syntactic sugar, notation, and declaration structure required to faithfully reconstruct compilable Lean contexts.

Lastly, to allow for iterative compilation attempts, we introduce the Lean-Compilation tool that provides the model a clean local Lean environment in which to test the compilation of the generated context for each benchmark problem. The tool returns the stack trace of failed compilation attempts to improve the agent's ability to debug failures.

The final output of this process is two-fold. First, we construct an agentic pipeline that faithfully retrieves minimal, compilable contexts from a large, custom Lean project. Second, we generate \DATANAME{}, a set of textbook-syntax exercises with minimal, fully compilable Lean context that preserves the intent of the original formalization. This benchmark enables a controlled evaluation of model performance under a genuine out-of-distribution formal setting, while avoiding confounding factors arising from missing context or compilation failures.

\subsection{\DATANAMEMathlib}

To isolate the effect of definitional frameworks from the mathematical difficulty of the problems, we translate from \DATANAME{} statements, written in Tao's definitional framework with in-context local definitions, into statements that use standard MathLib definitions, which we title \DATANAMEMathlib{}. These paired problems enable an in-distribution evaluation of existing public provers under the MathLib definitional framework. To perform this translation at scale, we build an automated pipeline and followed by expert manual verification of the resulting MathLib statements to ensure mathematical equivalence with their Tao counterparts.

Constructing a MathLib counterpart for each exercise in \DATANAME{} introduces a distinct but related set of technical challenges. In particular, such a reformulation must simultaneously satisfy two constraints: (i) the resulting statement must be syntactically valid and compilable within MathLib, and (ii) it must remain mathematically faithful to the original textbook formulation.

Naïve, end-to-end translation of textbook formalizations into MathLib using frontier language models proves unreliable. In early experiments, we attempted direct translation using GPT-5.1 with prompt engineering and manual inspection. These attempts yielded poor results: roughly 60\% of outputs were syntactically valid Lean code, and a substantially smaller fraction were mathematically equivalent to the original statements. Inspection of failure cases revealed that most errors stemmed from mismatches between Tao's local definitions and their MathLib counterparts. In particular, the model frequently selected incorrect definitions or employed non-canonical usage patterns, leading to both compilation failures and subtle semantic deviations.

\begin{figure*}[!t]
  \centering
  \includegraphics[width=0.98\textwidth]{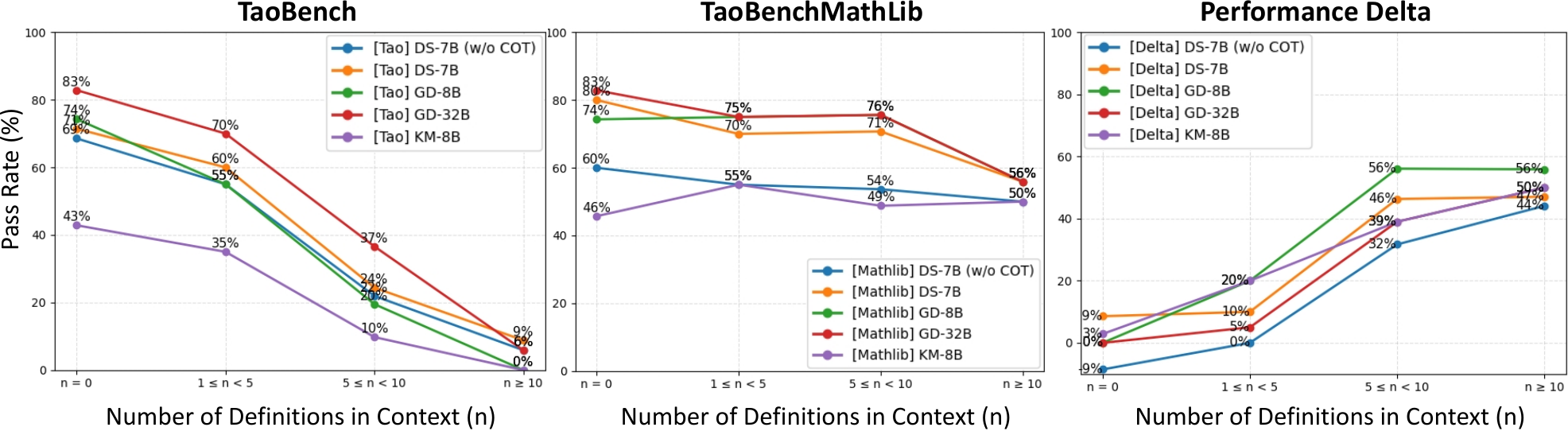}
  \caption{
  Model Performance vs. Definitions in Context. Performance on \DATANAME{} and \DATANAMEMathlib{} as a function of the number of in-context definitions. The performance delta is defined as performance~(\DATANAME{}) - performance~(\DATANAMEMathlib{}). DS, GD, and KM indicate DeepSeek-Prover-V2, Goedel-Prover-V2, and Kimina-Prover, respectively.
  }
  \label{fig:context_length_figure}
  \vspace{-4mm}
\end{figure*}

These observations indicate that accurate reformulation into MathLib requires grounding the model in MathLib's canonical definitions and usage conventions. To this end, we enable web-search to allow the model to retrieve authoritative documentation and examples for core MathLib constructs. With retrieval enabled, syntactic correctness improves substantially: under this configuration, approximately 90\% of generated statements compile successfully. This improvement suggests that access to curated MathLib usage patterns is critical for reliable translation, motivating the design of our translation pipeline.

Our translation pipeline consists of two stages: a rewriting stage and an equivalence-checking stage. In the rewriting stage, we prompt GPT-5.1 with web search enabled to reformulate the Tao-style statement into a MathLib-compatible version. We explicitly restrict the model from introducing new local definitions or notation and require adherence to a fixed MathLib header, ensuring consistency across translated problems. The resulting statement is then passed to the Lean compiler, and we employ an iterative refinement loop in which compiler error messages are used to resolve syntactic and typing errors until the statement compiles successfully.

Compilation alone, however, does not guarantee semantic faithfulness. To address this, we introduce an equivalence-checking stage. Using the JiXia tool, we extract the target proof states for both the Tao and MathLib formulations, grounding the comparison in their respective proof obligations. We then prompt GPT-5.1, again with web search enabled, to assess whether the two formulations are mathematically equivalent given their statements and target states. Candidates that fail this equivalence check are discarded, and the pipeline is restarted to regenerate a valid reformulation. Appendix~\ref{appendix:Mathlib-Translation-Pipeline} reports detailed statistics on compilation failures and equivalence-check rejections encountered during this process.

This pipeline generates \DATANAMEMathlib{}, a MathLib-based reformulation of \DATANAME{} that preserves the mathematical intent of the original textbook exercises while remaining fully compatible with MathLib’s definitional framework. Together with \DATANAME{}, this paired construction enables controlled evaluation of automated theorem provers under definitional shift, isolating the effects of formal framework differences from confounding factors such as compilation failures or semantic drift.

\subsubsection{Expert Manual Verification}
To verify the quality of both versions of the benchmark, we had expert human annotators provide determinations on whether the \DATANAME{} faithfully retrieved the context for each exercise, and whether each MathLib formalization was mathematically equivalent to its Tao counterpart. Each annotator had completed an equivalent \textit{Analysis I} course and had experience in formal proving in Lean. Issues identified by annotators in context retrieval helped us construct a final agentic context retrieval pipeline that produces correct context for all 150 problems initially chosen to be in \DATANAME{}. Human annotators identified 33 often subtle issues with the MathLib translations and provided corrected versions of the problems.
The full protocol for this is provided in Appendix~\ref{appendix:Expert-Manual-Annotation}.

\begin{table}[!t]
\centering

\caption{Model performance on \DATANAME{} with and without the retrieved context as part of the input.}

\normalsize

\renewcommand{\arraystretch}{1}
\begin{tabular}{
p{0.22\textwidth}
>{\centering\arraybackslash}p{0.10\textwidth}
>{\centering\arraybackslash}p{0.10\textwidth}
}
\toprule
\multirow{2}{*}{\textbf{Model}} &
\multicolumn{2}{c}{\DATANAME{}} \\
\cmidrule(lr){2-3}
& \textbf{Pass@128} & \textbf{(\%)} \\
\midrule

DeepSeek-7B w/o COT                    & 57/150  & 38.00  \\
- w/o Context                          & 41/150  & 27.33  \\

\noalign{\vskip 0.5mm}
\cdashline{1-3}
\noalign{\vskip 0.5mm}

DeepSeek-Prover-V2-7B                  & 62/150  & 41.33  \\
- w/o Context                          & 44/150  & 29.33  \\

\noalign{\vskip 0.5mm}
\cdashline{1-3}
\noalign{\vskip 0.5mm}

Goedel-Prover-V2-8B                    & 56/150  & 37.33  \\
- w/o Context                          & 41/150  & 27.33  \\

\noalign{\vskip 0.5mm}
\cdashline{1-3}
\noalign{\vskip 0.5mm}

Goedel-Prover-V2-32B                   & 74/150  & 49.33  \\
- w/o Context                          & 46/150  & 30.00  \\

\noalign{\vskip 0.5mm}
\cdashline{1-3}
\noalign{\vskip 0.5mm}

Kimina-Prover-8B                       & 33/150  & 22.00  \\
- w/o Context                          & 30/150  & 20.00  \\



\bottomrule
\end{tabular}
\renewcommand{\arraystretch}{1.0}

\label{tab:analysis_noContext}
\vspace{-3mm}
\end{table}

\begin{table*}[!t]
\centering

\caption{Model Performance on \DATANAME{} and  \DATANAMEMathlib{} for frontier foundation models.}

\normalsize

\renewcommand{\arraystretch}{1}

\begin{tabular}{
p{0.32\textwidth}
>{\centering\arraybackslash}p{0.12\textwidth}
>{\centering\arraybackslash}p{0.10\textwidth}
>{\centering\arraybackslash}p{0.12\textwidth}
>{\centering\arraybackslash}p{0.10\textwidth}
}
\toprule
\multirow{2}{*}{\textbf{Model}} &
\multicolumn{2}{c}{\DATANAME{}} &
\multicolumn{2}{c}{\DATANAMEMathlib{}} \\
\cmidrule(lr){2-3} \cmidrule(lr){4-5}
& \textbf{Pass@8} & \textbf{(\%)} & \textbf{Pass@8} & \textbf{(\%)} \\
\midrule

GPT-5.1~\cite{GPT5_1}  & 78/150  & 52.00 & 97/150 & 64.67 \\

Gemini 3 Pro~\cite{Gemini_3_Pro}  &  69/150  &  46.00  &  91/150  &  60.67 \\

DeepSeek-V3.2~\cite{DeepSeek_v3_2}  & 71/150  & 47.33 & 105/150 & 70.00 \\

\bottomrule
\end{tabular}

\renewcommand{\arraystretch}{1.0}

\label{tab:tao-mathlib-results-proprietarymodel}
\vspace{-2mm}
\end{table*}

\section{Experiments}
We conduct comprehensive experiments on our benchmarks. 
Section~\ref{subsec:experiment_setup} describes  experimental setup, and Section~\ref{subsec:main_results} presents our main results and analysis.

\subsection{Experimental Setup}
\label{subsec:experiment_setup}

We evaluate four public SOTA ATP models: DeepSeek-Prover-v2-7B~\cite{deepseek_prover_v2}, Goedel-Prover-v2-8B~\cite{goedel_prover_v2}, Goedel-Prover-v2-32B~\cite{goedel_prover_v2}, and Kimina-Prover-8B~\cite{kimina_prover}. We use default inference and decoding settings with a temperature of 1.0 and reasoning chains of up to 8192 tokens. We report pass@128 accuracy as the primary metric for each model.

\subsection{Main Results}
\label{subsec:main_results}

Table~\ref{tab:tao-mathlib-results} reports pass@128 results on \DATANAME{} and \DATANAMEMathlib. Although the two versions of the benchmarks consist of mathematically equivalent problem sets, all evaluated provers perform substantially worse on the Tao formulations than on the MathLib formulations. This consistent degradation highlights the difficulty current provers face under a shift in definitional frameworks.

The performance gap persists across model families, parameter scales, and inference configurations, indicating that it reflects a shared limitation rather than model-specific deficiencies. On \DATANAME{}, overall performance remains low: the strongest result is achieved by Goedel-Prover-V2-32B at 49.33\%, while the best 7B-8B model, DeepSeek-Prover-V2-7B, reaches 41.33\%. In contrast, on \DATANAMEMathlib, most models achieve accuracies above 65\%, with several exceeding 70\%. This uniform gap suggests that current public ATPs generalize poorly outside the MathLib definitional framework on which they are primarily trained and evaluated.

Scaling improves performance on both benchmarks but does not eliminate the discrepancy. Goedel-Prover-V2-32B improves by 12 percentage points over Goedel-Prover-V2-8B on \DATANAME{}, compared to only a 2-point improvement on \DATANAMEMathlib{}. While increasing model scale provides modest robustness to definitional shift, even the strongest model falls well short of matching its performance on \DATANAMEMathlib{}. 
%
\DATANAME{} therefore exposes a clear gap between in-distribution benchmark success and robustness to definitional variation.

\begin{figure*}[t]
  \centering
  \includegraphics[width=1\textwidth]{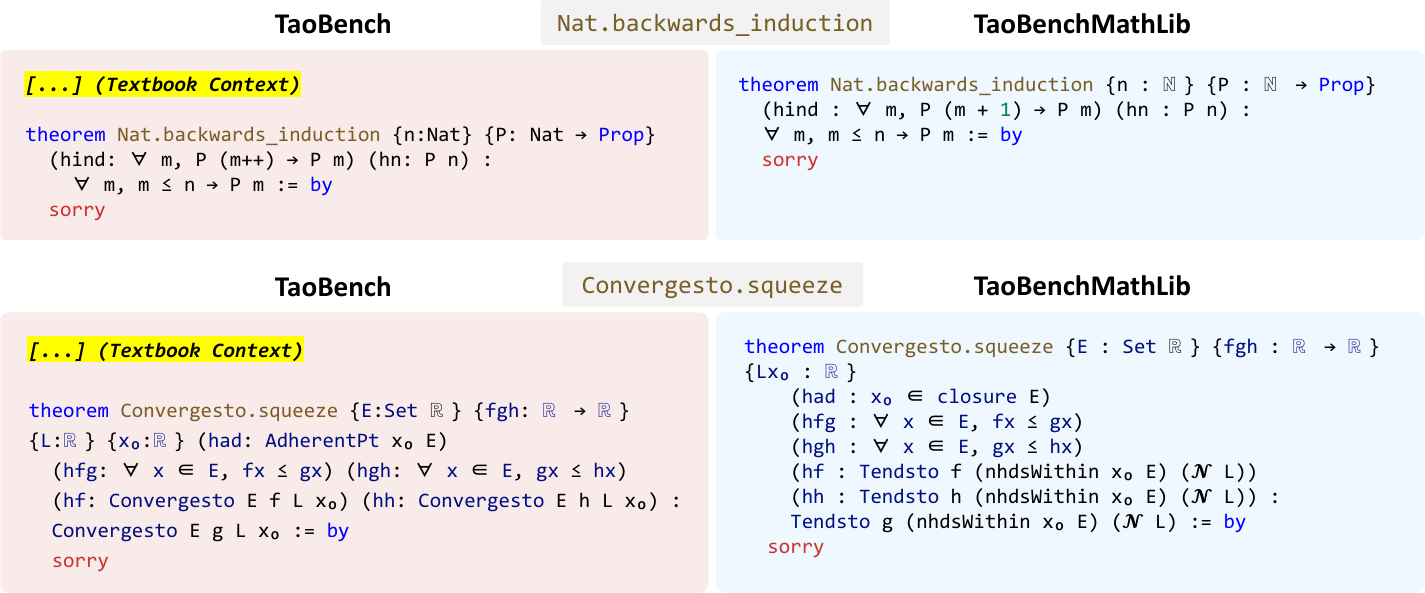}
  \caption{
 Case studies: Nat.backwards\_induction and Convergesto.squeeze. The textbook context for TaoBench is omitted due to space limitations. The full statements of the two exercises are shown in Figures~\ref{fig:case_study_full_1} and~\ref{fig:case_study_full_2}.
  }
  \label{fig:case_study_1}
  \vspace{-4mm}
\end{figure*}

\section{Analysis}

\subsection{Context Length}
\label{subsec:context_length}

\begin{table}[t]

\centering

\caption{Counts of data samples across ranges of different numbers of local definitions.}

\renewcommand{\arraystretch}{1}
\resizebox{1\linewidth}{!}{
\begin{tabular}{
  >{\centering\arraybackslash}p{0.17\textwidth} 
  >{\centering\arraybackslash}p{0.04\textwidth} 
  >{\centering\arraybackslash}p{0.05\textwidth}
  >{\centering\arraybackslash}p{0.06\textwidth} 
  >{\centering\arraybackslash}p{0.04\textwidth} 
}

\toprule

\# Local Definitions & 0 & [1, 5) & [5, 10) & 10 +  \\

\midrule

Data Count & 35 & 40 & 41 & 34  \\

\bottomrule
\end{tabular}
}
\label{Table:num_definitions}

\vspace{-4mm}

\end{table}

We analyze how model performance is impacted by the number of local definitions provided in each context. 
Problems in \DATANAME{} contain the minimal local environment (bespoke definitions, notation, and supporting lemmas) to ensure the statement compiles, whereas \DATANAMEMathlib{} is evaluated in the standard \texttt{import Mathlib} environment and does not contain additional local definitions in the prompt. 
Figure~\ref{fig:context_length_figure} divides the benchmark into four buckets based on the number of locally introduced definitions required to state each problem: $n = 0$, $1 \le n < 5$, $5 \le n < 10$, and $n \ge 10$. We include the number of problems in each bucket in Table~\ref{Table:num_definitions}.

A consistent and sharp divergence emerges between performance on the Tao and MathLib formulations as context length increases. Model performance on the \DATANAMEMathlib{} version decays slowly as the number of definitions in the \DATANAME{} counterpart increases. Averaged across models, pass@128 decreases only from 68.10\% at $n=0$ to 53.43\% at $n \ge 10$. In contrast, \DATANAME{} exhibits a significant degradation as context length increases. Average pass@128 drops from 67.14\% at $n=0$ to just 6.37\% at $n \ge 10$. This collapse is consistent across all evaluated models, and scales to larger models. 

We also observe that the performance delta between \DATANAME{} and \DATANAMEMathlib{} performance increases with context length. At $n=0$, the average delta across models is near zero, which confirms that Tao problems without local definitions are not inherently harder than their MathLib counterparts. However, when the number of in-context definitions increases to $5 \le n < 10$, the average delta exceeds 40 percentage points, and for $n \ge 10$, it approaches or exceeds 50 percentage points for all models. This pattern suggests that the performance difference is 
driven by an inability to effectively integrate and reason over unfamiliar definitions introduced in context.

Formal theorem provers are routinely expected to operate in novel definitional environments, where new abstractions must be introduced and used within a local context. Our results show that, despite having necessary definitions in context, current models struggle to apply them when those definitions shift from the MathLib distribution. This suggests that fine-tuning improves performance within a fixed formal ecosystem, but does not necessarily generalize to unseen definitional frameworks, exposing a fundamental limitation of current ATP training paradigms.

\subsection{Context Dependency}
\label{subsec:context_vs_no_context}

To provide insight into the effect of the context provided in the \DATANAME{} versions of the problems, we evaluate model performance when local definitions are removed from the prompt. In this setting, models are presented with the problem statements from \DATANAME{} without any accompanying local environment, even when the statements reference custom definitions introduced in the textbook formalization.

As shown in Table~\ref{tab:analysis_noContext}, across all models, removing contextual information leads to a consistent, but relatively modest drop in performance compared to the full-context Tao setting. 
 A manual review of the problems that models solved without context shows that the associated set of references either contained no additional statements (i.e., context length of zero) or referenced a small set of statements that were similar to or in the standard MathLib definitional framework (ex., the rational numbers). Further manual review of the proofs associated with these problems found that the problems were often made easier by using MathLib constructions to circumvent the crux of the exercise (see \ref{subsec:case_studies} for examples). 

Taken with the context-length analysis, these results indicate that current formal theorem provers do not reliably benefit from in-context definitional information unless it conforms closely to their training distribution. The presence of explicit local definitions is therefore not inherently helpful and can be actively harmful, highlighting a limitation in models’ ability to perform in-context definitional grounding.

\subsection{Frontier Models}
\label{subsec:frontier_models}

We also evaluate frontier foundation models in Table~\ref{tab:tao-mathlib-results-proprietarymodel} to provide context for the performance of SOTA ATP models. Although these models are not explicitly trained for formal theorem proving, they are effective at using in-context information to solve unfamiliar problems. This comparison allows us to assess whether the Tao vs. MathLib performance gap reflects issues specific to ATP specialization or a broader difficulty posed by the Tao definitional framework.

As shown in Tables~\ref{tab:tao-mathlib-results} and \ref{tab:tao-mathlib-results-proprietarymodel}, frontier models underperform both Goedel-Prover-V2-32B and several 7B parameter ATP models on the MathLib problem formulations, which indicates that ATP models excel in definitional frameworks in which they are trained. However, while the performances of  the frontier models do degrade on the Tao problem formulations, they achieve performances comparable to  -- or even in excess of -- Goedel-Prover-V2-32B. These results indicate that the relative advantage of frontier models on Tao formulations is driven by a superior capability to leverage in-context examples rather than overall stronger proving capabilities.

\subsection{Case Studies}
\label{subsec:case_studies}

We present two representative instances taken from model performance on complementary problem pairs: one where the prover succeeds on the MathLib formulation but not on the Tao formulation, and one showing the reverse pattern.

\textbf{\texttt{Nat.backwards\_induction}}:
As shown in Figure~\ref{fig:case_study_1}, this pair highlights how prover performance can depend on access to MathLib's arithmetic and order-theoretic infrastructure. Empirically, for Goedel-Prover-V2-32B, the \DATANAME{} version admits 0/128 verified proofs, while the \DATANAMEMathlib{} version admits 64/128 verified proofs. 

In the \DATANAME{} prompt, the prover must reason over a bespoke inductive \texttt{Nat} with addition defined by primitive recursion and order defined existentially by \(m \le n,\ \exists a,\, n = m + a\). A backwards-induction argument, therefore, requires reconstructing supporting algebraic and order-theoretic lemmas locally. In the \DATANAMEMathlib{} prompt, these obligations can be met using standard lemmas and automation. The instance-level performance split (0/128 vs.\ 64/128 for Goedel-32B) corresponds with brittleness to definitional relocation rather than increased mathematical difficulty.

\textbf{\texttt{Convergesto.squeeze}}:
As shown in Figure~\ref{fig:case_study_1}, this pair highlights a representational shift that is common throughout analysis formalizations: an explicit $\varepsilon$--$\delta$ definition of convergence versus a filter-based \texttt{Tendsto} statement. Empirically, for DeepSeek-Prover-V2-7B, the \DATANAME{} version admits 2/128 verified proofs, while the \DATANAMEMathlib version admits 0/128 verified proofs. 

In the \DATANAME{} prompt, \texttt{Convergesto} is defined directly via \texttt{CloseNear}: for each $\varepsilon>0$, one produces a $\delta>0$ so that \(|f(x)-L|<\varepsilon\) holds on a punctured $\delta$-neighborhood intersected with $E$. A successful proof proceeds by extracting radii $\delta_f,\delta_h$ from the hypotheses for $f$ and $h$, choosing $\delta=\min(\delta_f,\delta_h)$, and then deriving the bound for $g$ by combining the pointwise inequalities $f\le g\le h$ with the two-sided inequalities obtained from rewriting \texttt{abs\_sub\_lt\_iff}. In contrast, the \DATANAMEMathlib{} version expresses the same squeeze argument as a filter limit \texttt{Tendsto g (nhdsWithin \(x_0\) E) \((\mathcal{N}\, L)\)}. Closing the goal in that setting requires translating pointwise bounds on $E$ into \texttt{Eventually} statements in \texttt{nhdsWithin} and then applying an appropriate squeeze principle at the level of filters. The observed asymmetry suggests that, for this model, proof search is more reliable when convergence is presented in an explicit, locally unfolded form than when it is mediated through MathLib's filter abstraction, even when the underlying mathematics is unchanged.

\section{Related Works}

\subsection{Formal Mathematical Benchmarks}
Early benchmarks for ATP models involved large-scale extraction of theorems and complementary proofs from MathLib~\cite{leanDojo}. However, the standard for benchmarking formal reasoning capabilities has become high-school-level competition math exam problems, such as the International Math Olympiad
~\cite{Trinh2024_OlympiadGeometry, LuongLockhart2025_GeminiBlog, OpenAI2025_tweet, Yu2025_FormalMATH, Hubert2025OlympiadRL}. Competition-style benchmarks such as \textsc{miniF2F}~\cite{miniF2F}, AIME~\cite{AIME_dataset}, and  PutnamBench~\cite{putnamBench} aggregate competition problems from past years and are widely used as standard evaluation targets for end-to-end proving performance. Other datasets focus on translating from informal language to formal languages: ProofNet~\cite{proofNet}, the HERALD dataset~\cite{herald}, and Lean Workbook~\cite{lean_workbook} provide natural-language statements (and often proofs) with mathematically equivalent formalizations, enabling evaluation of both auto-formalization and proof search. More recent efforts expand topical coverage 
beyond contest mathematics~\cite{combiBench, FATE, deepseek_prover_v2}. 


\subsection{Automated Theorem Proving}
Automated theorem proving in Lean has evolved rapidly, with earlier approaches leveraging inference-time strategies \cite{leanDojo, lean_STaR, BFS_prover_V2} and more recent approaches using LLMs for end-to-end generation \cite{deepseek_prover_v2, goedel_prover_v2, kimina_prover}.
Inference-time strategies such as best-first or expert-iteration search established strong performance on standard benchmarks~\cite{BFS_prover_V2}. Subsequent approaches that incorporated synthetic data generation and reinforcement learning further improved the state-of-the-art \cite{STP}.
The DeepSeek-Prover-V2 \cite{deepseek_prover_v2}, Kimina-Prover \cite{kimina_prover}, and Goedel-Prover-V2 model \cite{goedel_prover_v2} families adopt iterative self-improvement loops that bootstrap from successful proofs and sub-goal decompositions. However, these pipelines rely largely on MathLib formalizations, raising open questions about robustness and generalization under shifts in underlying definitions and formal representations.


\section{Conclusion}

Our work identifies a concrete limitation of current public LLM-based ATP models. We observe that strong performance on Mathlib-formulated problems does not reliably transfer to definitionally equivalent formalizations outside the MathLib framework. Across model families and scales, accuracy degrades substantially when core mathematical concepts are reintroduced through alternative definitions. 
Our analysis of the impact of context features and manual evaluation of the results indicate that this gap is due more to sensitivity to novel formulations  in \DATANAME{} than to increased mathematical difficulty over \DATANAMEMathlib. This poses an acute problem because much of exploratory mathematics takes place in novel definitional frameworks that are distinct from MathLib. 
By isolating the effects of changing definitional frameworks, \DATANAME{} and \DATANAMEMathlib{} show that existing ATP benchmarks conflate general mathematical ability with familiarity with a specific definitional framework. 
We believe that \DATANAME{}, \DATANAMEMathlib, and our agentic pipelines will provide useful tools for future research into public ATP models that can be useful to research mathematics.


\clearpage

\subsubsection*{Impact Statements}

This paper presents work whose goal is to advance the field of machine learning. There are many potential societal consequences of our work, none of which we feel must be specifically highlighted here.

\section{Acknowledgements}
We would like to thank Dr. Terence Tao for providing assistance and insight into the construction of his \textit{Analysis I} textbook. Additionally, this work was partially supported by NSF 2312501, 2106859, NIH 1U54HG012517-01, U54DK097771, U54OD036472, OT2OD038003, Amazon, NEC, and Optum Labs.

\bibliographystyle{style_sheets/icml2026}
\bibliography{custom}

\newpage
\appendix

\clearpage

\section{Appendix}

\subsection{Expert Manual Annotation of \DATANAMEMathlib{}}
\label{appendix:Expert-Manual-Annotation}
We verified the fidelity of the MathLib equivalent problems translated from the textbook versions via both manual expert reviews and LLM-as-evaluator passes. The reviewer qualifications were completion of an equivalent Analysis I course and experience in Lean programming. We provided both the Tao and MathLib versions of the data to the annotators and asked the annotators to gauge mathematical equivalence according to the grades outline in Table~\ref{tab:label_definition}. Annotators were asked to provide a manual translation of the Tao example in MathLib syntax if they identified MathLib translations that were graded as a "C" or a "D". The interface for human annotation is shown in Figure~\ref{fig:Expert_Annotation_Interface}. We also applied an additional annotation pass by GPT-5-Pro: we provide the model with the annotation guidelines and in-context examples of corrections made by human annotators and query the model to grade the translation and provide a re-translation if the translation grade is a "C" or a "D". One annotator manually reviewed the grades given by the LLM and proposed corrections to produce the final version of the benchmark.

\paragraph{Judgment Labels}
Our labeling scheme is motivated by an important practical consideration: some definitions in Tao's \textit{Analysis I} differ from their MathLib counterparts. In the Lean formalization of \textit{Analysis I}, Tao generally favors faithful translation over more idiomatic Lean formulations when the two diverge~\cite{tao_analysis_github}. Consequently, certain textbook exercises do not admit a strictly equivalent statement in MathLib. We therefore instructed judges to evaluate the core intent of each exercise and the corresponding proof strategy across the two versions. If these are essentially preserved, we regard the pair as acceptable even when the MathLib statement differs slightly from the Tao statement. Based on these considerations, we define four judgment labels, summarized in Table~\ref{tab:label_definition}.

\paragraph{Judgment Process}
For each exercise pair, judges first identified the core intent of the two versions and considered whether their proof strategies align. If the intent and proof strategy were essentially the same, the instance was labeled A or B; otherwise, it was labeled C or D. For label B instances, judges were required to explicitly describe the differences between the two versions. For labels C and D, judges were required to (i) describe the differences and (ii) provide a corrected MathLib version of the exercise.
Each judge evaluated 70 data points. To measure inter-judge agreement, 30 of the 70 data points assigned to each judge were evaluated by all three judges.

\paragraph{Annotation Pipeline Results}
Based on the manual judgments, 108 data points were assigned label A and 31 were assigned label B, which is expected given the inherent definitional differences between Tao's textbook and MathLib. We identified 11 MathLib-version data points that required correction; these were manually revised by the judges. We computed Gwet's AC2 to quantify inter-judge agreement among the three judges and obtained a score of 0.89, indicating high reliability.

\begin{table}[!t]
\centering

\caption{Definitions of four judgment labels.}

\small
\renewcommand{\arraystretch}{1.2}
\begin{tabular}{p{0.12\linewidth} p{0.82\linewidth}}
\midrule
\textbf{Label} & \textbf{Description} \\
\midrule


A &
\textbf{Fully Equivalent.} The two versions express the same mathematical statement. They are logically equivalent under standard mathematical interpretations. \\
\midrule
B &
\textbf{Nearly Equivalent.} The exercise's intent and proof strategy remain essentially the same. However, due to definitional differences between Tao's textbook and Mathlib, the Mathlib formulation is slightly different from the corresponding Tao statement. \\
\midrule
C &
\textbf{Minor Differences.} The exercise's intent and proof strategy are not preserved across the two versions, but the discrepancy is minor and requires only small adjustments. \\
\midrule
D  &
\textbf{Major Differences.} The exercise's intent and proof strategy are not preserved across the two versions, and there is a substantial mismatch in meaning. \\

\midrule
\end{tabular}

\label{tab:label_definition}
\end{table}


\subsection{Data Distribution}

\begin{figure}[h]
  \centering
  \includegraphics[width=0.5\textwidth]{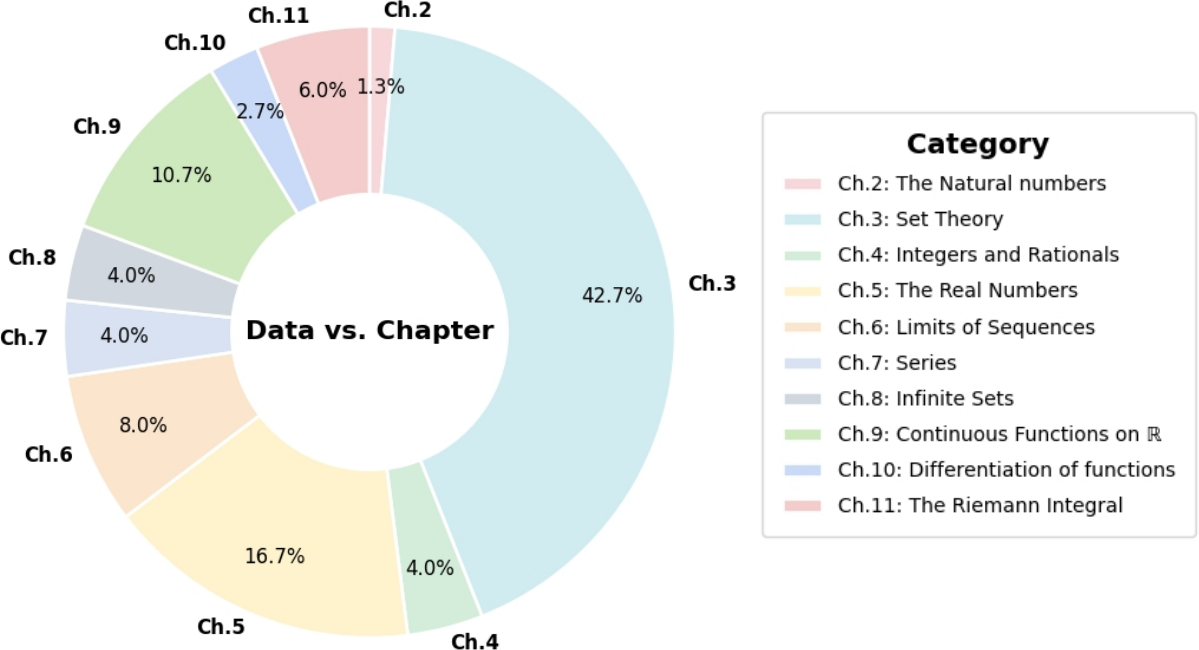}
  \caption{
  Data Distribution by Chapter.}
  \label{fig:PieChart_Chapter}
\end{figure}

Our benchmark comprises ten chapters (Chapters 2-11) and sixty-five subsections that follow a standard progression from foundational material (natural numbers, set theory, and number systems) to core real analysis topics, including limits, series, continuity, differentiation, and the Riemann integral. Figure~\ref{fig:PieChart_Chapter} summarizes the chapter-level composition.

\begin{table*}[!t]
\centering

\caption{Tree-Search Based Model Performance on \DATANAME{} and \DATANAMEMathlib{}. Results are reported as pass rates on 150 problems. Models without reported results for a given setting are marked as "--". }

\normalsize

\renewcommand{\arraystretch}{1}

\begin{tabular}{
p{0.32\textwidth}
>{\centering\arraybackslash}p{0.14\textwidth}
>{\centering\arraybackslash}p{0.10\textwidth}
>{\centering\arraybackslash}p{0.14\textwidth}
>{\centering\arraybackslash}p{0.10\textwidth}
}
\toprule
\multirow{3}{*}{\textbf{Model}} &
\multicolumn{2}{c}{\DATANAME{}} &
\multicolumn{2}{c}{\DATANAMEMathlib{}} \\
\cmidrule(lr){2-3} \cmidrule(lr){4-5}
& \textbf{Tactic Budget} & \textbf{Pass Rate} & \textbf{Tactic Budget} & \textbf{Pass Rate} \\
\cmidrule(lr){2-3} \cmidrule(lr){4-5}
& 128$\times$2$\times$600 & \textbf{(\%)} & 128$\times$2$\times$600 & \textbf{(\%)} \\
\midrule

BFS-Prover-V2-7B~\cite{BFS_prover_V2}                       & --      & --    & 103/150 & 68.67 \\

BFS-Prover-V2-32B~\cite{BFS_prover_V2}                      & --      & --    & 105/150 & 70.00 \\

\bottomrule
\end{tabular}

\renewcommand{\arraystretch}{1.0}
\vspace{2mm}

\label{tab:tao-mathlib-results-BFS-Prover}

\vspace{-2mm}
\end{table*}

\subsection{Model Performance by Chapter}

\begin{figure*}[h]
  \centering
  \includegraphics[width=0.98\textwidth]{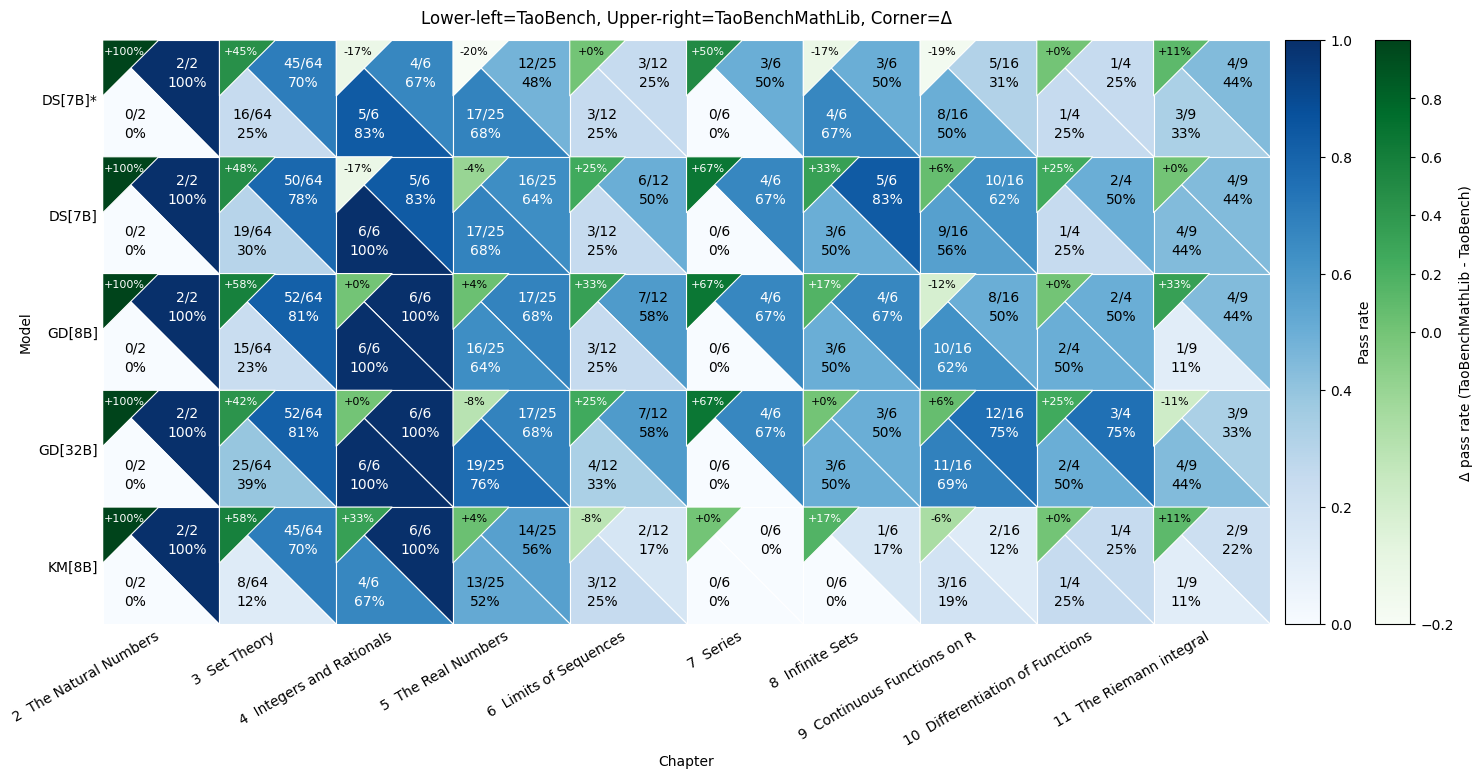}
  \caption{
  Model Performance by Chapter. For each cell, accuracies on \DATANAME{} and \DATANAMEMathlib{} are shown in the upper-right and lower-left corners, respectively. The upper-left corner shows the difference (\DATANAMEMathlib{} - \DATANAME{}). DS[7B]* indicates DeepSeek-Prover-V2-7B (w/o COT).
  }
  \label{fig:Performance_vs_Chapter}
\end{figure*}

\begin{table*}[h]
\centering

\caption{Model Performance by Chapter}

\normalsize

\renewcommand{\arraystretch}{1.03}
\begin{tabular}{
>{\centering\arraybackslash}p{0.05\textwidth}
>{\centering\arraybackslash}p{0.23\textwidth}
>{\centering\arraybackslash}p{0.05\textwidth}
>{\centering\arraybackslash}p{0.05\textwidth}
>{\centering\arraybackslash}p{0.05\textwidth}
>{\centering\arraybackslash}p{0.05\textwidth}
>{\centering\arraybackslash}p{0.05\textwidth}
>{\centering\arraybackslash}p{0.05\textwidth}
>{\centering\arraybackslash}p{0.05\textwidth}
>{\centering\arraybackslash}p{0.05\textwidth}
}
\toprule
\multicolumn{1}{c}{\textbf{Chapter}} &
\multicolumn{1}{c}{\textbf{Title}} &
\multicolumn{1}{c}{\textbf{\# Data}} &
\multicolumn{1}{c}{\textbf{DS\textsubscript{[7B]}\textsuperscript{*}}} &
\multicolumn{1}{c}{\textbf{DS\textsubscript{[7B]}}} &
\multicolumn{1}{c}{\textbf{GD\textsubscript{[8B]}}} &
\multicolumn{1}{c}{\textbf{GD\textsubscript{[32B]}}} &
\multicolumn{1}{c}{\textbf{KM\textsubscript{[8B]}}} &
\multicolumn{1}{c}{\textbf{GPT-5.1}} &
\multicolumn{1}{c}{\textbf{Average}} \\

\midrule

\multicolumn{9}{c}{\textit{\textbf{\DATANAME{}}}} \\

\midrule

2  & The Natural Numbers          & \emph{2}  & 0  & 0  & 0  & 0  & 0  & 0  & 0\%  \\
3  & Set Theory                   & \emph{64} & 16 & 19 & 15 & 25 & 8  & \underline{41} & 32\%  \\
4  & Integers and Rationals       & \emph{6}  & 5  & \underline{6}  & \underline{6}  & \underline{6}  & 4  & \underline{6}  & 92\%  \\
5  & The Real Numbers             & \emph{25} & 17 & 17 & 16 & \underline{19} & 13 & 13 & 63\%  \\
6  & Limits of Sequences          & \emph{12} & 3  & 3  & 3  & \underline{4}  & 3  & 3  & 26\%  \\
7  & Series                       & \emph{6}  & 0  & 0  & 0  & 0  & 0  & 0  & 0\%  \\
8  & Infinite Sets                & \emph{6}  & \underline{4}  & 3  & 3  & 3  & 0  & 3  & 44\%  \\
9  & Continuous Functions on R    & \emph{16} & 8  & 9  & 10 & \underline{11} & 3  & 9  & 52\%  \\
10 & Differentiation of Functions & \emph{4}  & 1  & 1  & \underline{2}  & \underline{2}  & 1  & 1  & 33\%  \\
11 & The Riemann integral         & \emph{9}  & 3  & \underline{4}  & 1  & \underline{4}  & 1  & 2  & 28\%  \\


\midrule

\multicolumn{9}{c}{\textit{\textbf{\DATANAMEMathlib{}}}} \\

\midrule

2  & The Natural Numbers           & \emph{2}  & 2  & 2   & 2   & 2   & 2   & 2  & 100\%  \\
3  & Set Theory                    & \emph{64} & 45 & 50  & 52  & 52  & 45  & 51 & 77\%   \\
4  & Integers and Rationals        & \emph{6}  & 4  & 5   & 6   & 6   & 6   & 6  & 92\%  \\
5  & The Real Numbers              & \emph{25} & 12 & 16  & 17  & 17  & 14  & 14 & 60\%  \\
6  & Limits of Sequences           & \emph{12} & 3  & 6   & 7   & 7   & 2   & 7  & 44\%   \\
7  & Series                        & \emph{6}  & 3  & 4   & 4   & 4   & 0   & 1  & 44\%   \\
8  & Infinite Sets                 & \emph{6}  & 3  & 5   & 4   & 3   & 1   & 4  & 56\%   \\
9  & Continuous Functions on R     & \emph{16} & 5  & 10  & 8   & 12  & 2   & 8  & 47\%   \\
10 & Differentiation of Functions  & \emph{4}  & 1  & 2   & 2   & 3   & 1   & 1  & 42\%  \\
11 & The Riemann integral          & \emph{9}  & 4  & 4   & 4   & 3   & 2   & 3  & 37\%  \\

\bottomrule
\end{tabular}
\renewcommand{\arraystretch}{1.0}

\label{tab:performance-chapter-all}
\end{table*}

\begin{figure*}[h]
  \centering
  \includegraphics[width=1\textwidth]{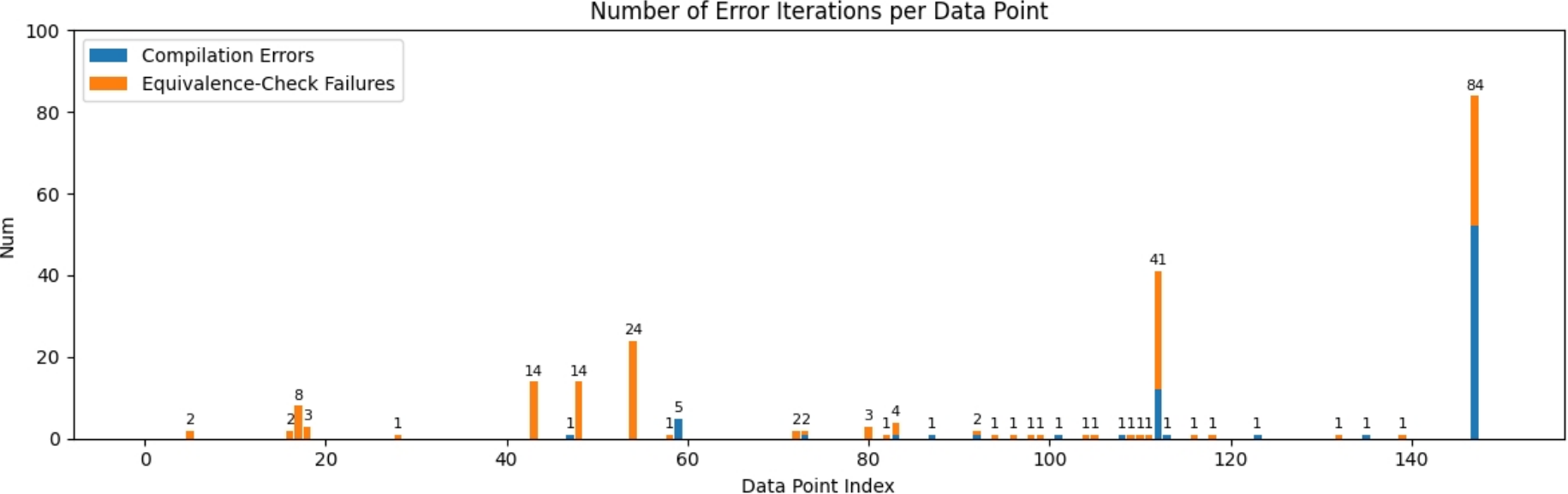}
  \caption{
  Number of Error Iterations per Data Point}
  \label{fig:Number_Error_Iterations}
\end{figure*}

Figure 7 breaks down pass rates by chapter and shows that the overall Tao-MathLib gap is not uniform: it is concentrated in a few chapters where the representation of core objects diverges most from MathLib. A large and consistent gap appears in Set Theory (Ch. 3), where all five models exhibit a substantial MathLib advantage of roughly 42-58 percentage points. Two pronounced failure modes occur in The Natural Numbers (Ch. 1) and Series (Ch. 7): for all five models, Tao accuracy falls to 0\%, while the MathLib translation remains solvable, suggesting that the bottleneck is not the underlying analysis content but rather sensitivity to the new definitional framework.

Outside these chapters, the picture is more mixed and often near parity. Several chapters show small deltas (e.g., Real Numbers (Ch. 5) and Continuous Functions (Ch. 9)), and in some cases the gap changes sign, indicating that Tao formulations are not systematically harder. Instead, performance depends on how closely the local definitions align with the models' MathLib-centric priors.

\begin{figure*}[h]
  \centering
  \includegraphics[width=1\textwidth]{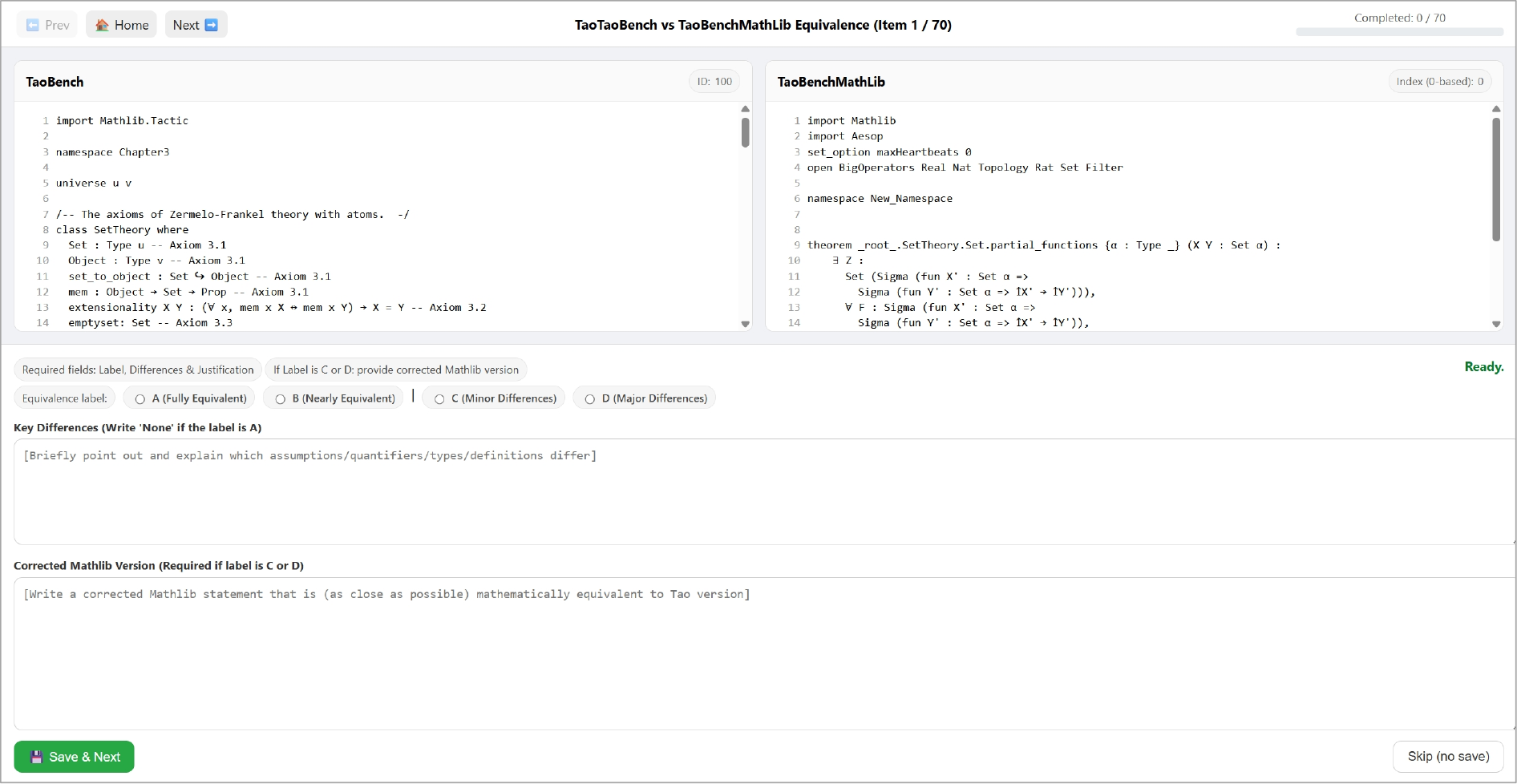}
  \caption{
  Human Annotation Interface}
  \label{fig:Expert_Annotation_Interface}
\end{figure*}

\begin{figure*}[h]
  \centering
  \includegraphics[width=1\textwidth]{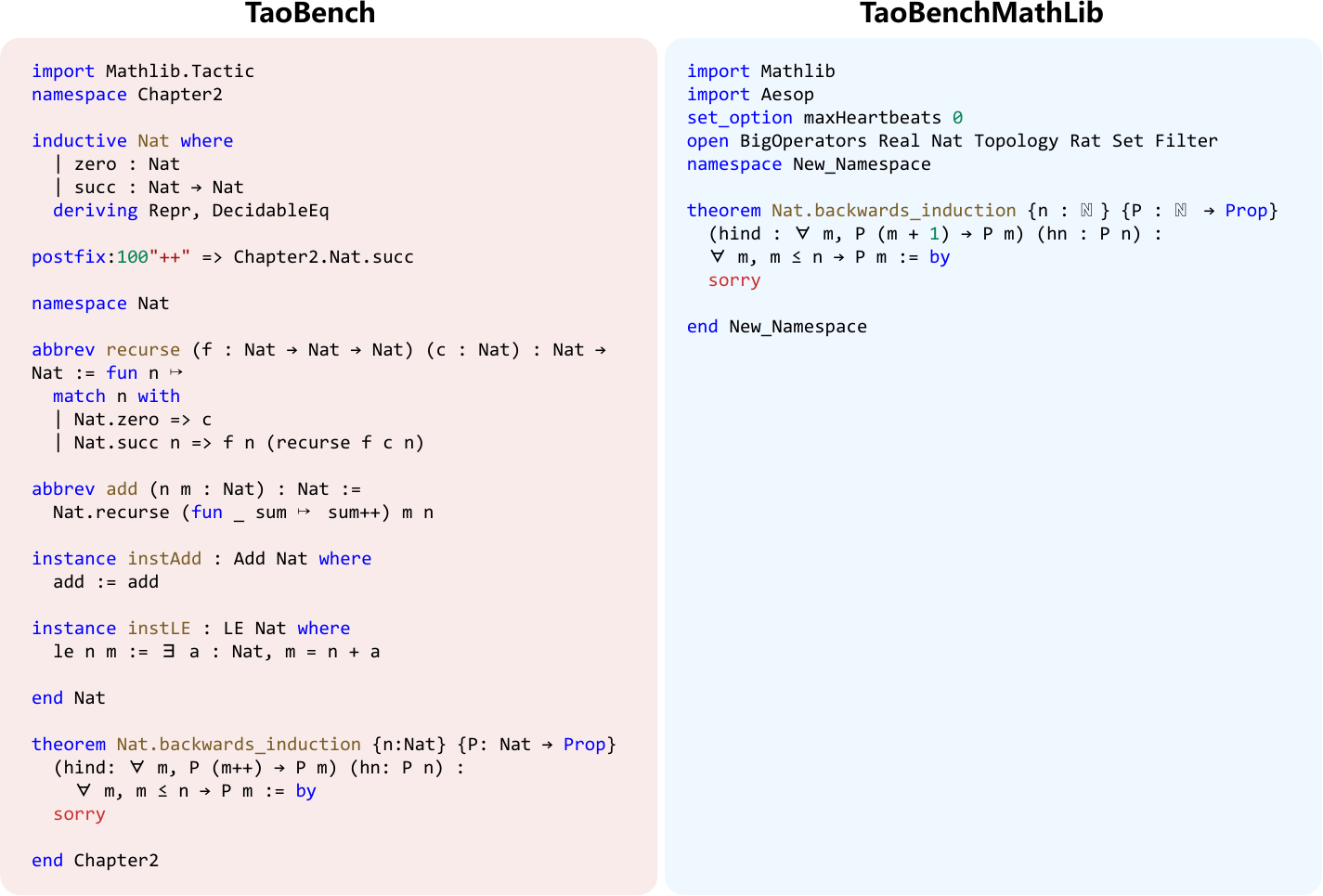}
  \caption{
  Case Study: Nat.backwards\_induction
  }
  \label{fig:case_study_full_1}
\end{figure*}

\begin{figure*}[h]
  \centering
  \includegraphics[width=1\textwidth]{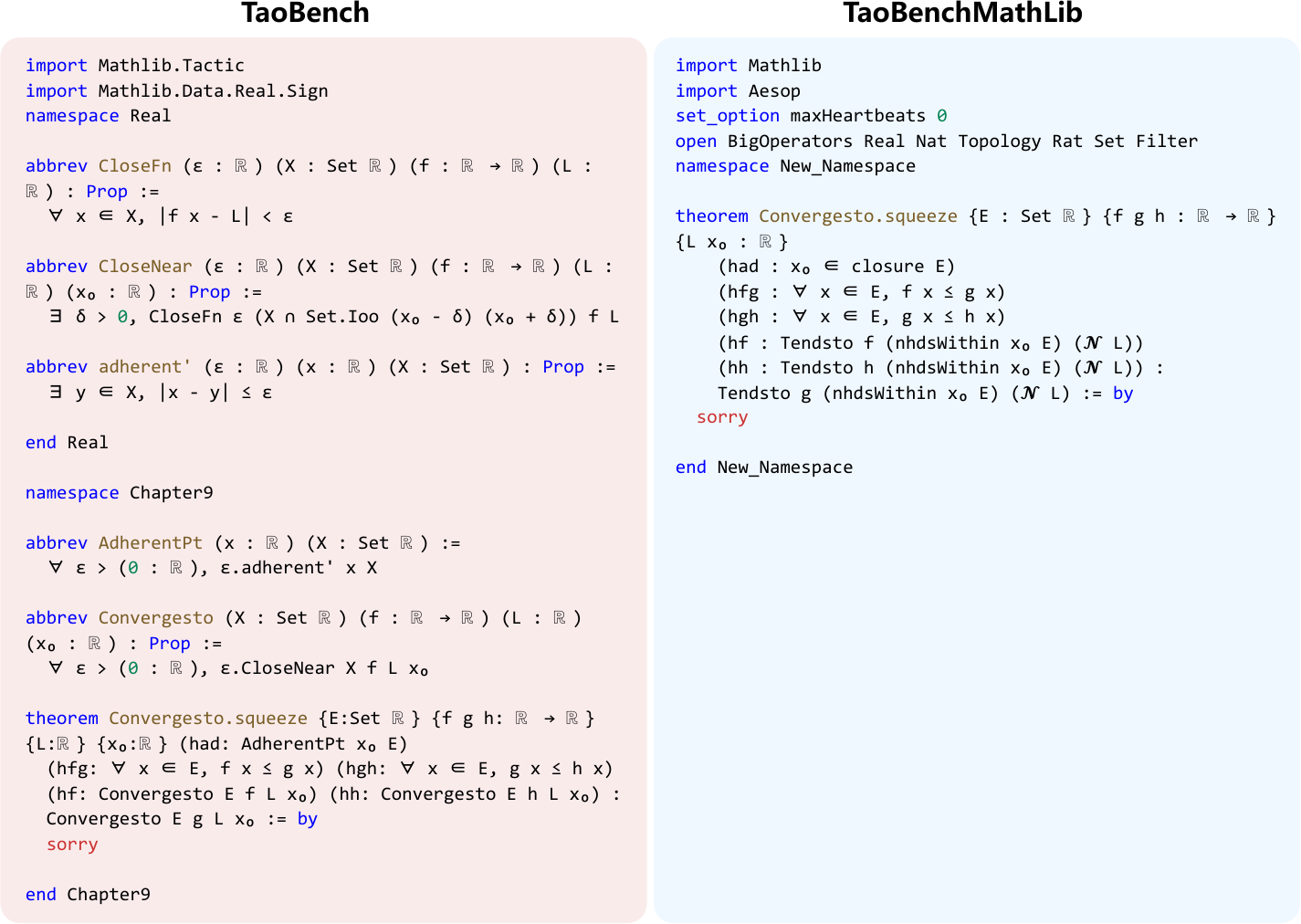}
  \caption{
  Case Study: Convergesto.squeeze
  }
  \label{fig:case_study_full_2}
\end{figure*}

\subsection{Additional Discussion}

\subsubsection{Tree-Search Based Model Evaluation}
\label{appendix:Discussion-BFS-Prover-Evaluation}

The evaluation results of a representative tree-search-based model, BFS-Prover, are shown in Figure~\ref{tab:tao-mathlib-results-BFS-Prover}. BFS-Prover is a stepwise prover. At each step, it generates the next tactic and interacts with LeanDojo to validate the tactic and obtain the next proof state. Our Tao-version exercises require Lean v4.23.0-rc2 or newer for compilation, but LeanDojo currently lacks support for these Lean versions. Therefore, we are unable to evaluate BFS-Prover on the Tao version of our benchmark.

For the evaluation on \DATANAMEMathlib{}, we adopt a tactic budget of 128 $\times$ 2 $\times$ 600, i.e., passes $\times$ expand\_k $\times$ max\_expansions. Concretely, we run up to 128 independent passes. In each pass, we expand at most 600 search nodes, and at each expansion we sample up to 2 tactic candidates from the model.

\subsubsection{MathLib Translation Pipeline}
\label{appendix:Mathlib-Translation-Pipeline}

During the rewriting stage, we forbid the MathLib version from introducing any new local definitions or notation, and restrict it to using the following fixed header:

\begin{minted}[breaklines,breakanywhere]{lean}
import Mathlib
import Aesop
set_option maxHeartbeats 0
open BigOperators Real Nat Topology Rat Set Filter
\end{minted}

For the rewriting and equivalence checking stages of our MathLib translation pipeline, we additionally count the number of compilation errors and equivalence-check failures for each data point, as shown in Figure~\ref{fig:Number_Error_Iterations}. Out of 150 data points, 30 encountered at least one error iteration. Across all error iterations, we observe 154 equivalence-check failures, which is substantially more than the 79 compilation errors. This suggests that ensuring mathematical equivalence is much harder than ensuring compilability during translation.

\subsection{System Prompts and Prompt Templates}
To aid in similar future efforts, we include the prompts used in the two pipelines detailed in this work. We provide (1) the final system prompt for the orchestrating agent in the textbook extraction pipeline(\ref{box:tao-System-Prompt}), (2) the prompt template used for each query in the extraction task(\ref{box:tao-Query-Prompt}), (3) the initial prompt for the rewriting stage of MathLib translation (\ref{box:tao-Initial-Rewriting-Prompt}), (4) the prompt for iterative compilation-error correction during rewriting (\ref{box:tao-Error-Correction-Prompt}), and (5) the prompt for the equivalence-checking stage of MathLib translation (\ref{box:tao-Equivalence-Checking-Prompt}).




\clearpage

\onecolumn
\subsubsection{Orchestrating Agent System Prompt for Textbook Extraction}
\label{box:tao-System-Prompt}
\begin{tcolorbox}[
    colframe=black, 
    colback=white, 
    coltitle=white, 
    colbacktitle=black, 
    title=Orchestrating Agent System Prompt for Textbook Extraction,
    fonttitle=\bfseries,
    boxrule=1pt,
    arc=2mm,    
    left=4mm, right=4mm, top=3mm, bottom=3mm, 
    enhanced, 
    fontupper=\small,
    breakable
]
{
\setlength{\parskip}{5pt}
You are a Lean compiler agent working inside an existing textbook project.

Your job is NOT to design new mathematics. Your job is to EXTRACT a faithful, self-contained Lean file so that one target theorem from a chapter can be compiled and typechecked in isolation, WITHOUT importing any `Analysis.Section\_*` modules.

There is no length limit. “Minimal” means dependency-minimal, not short.

--------------------------------

PRIMARY OBJECTIVES (IN ORDER)

--------------------------------

1. Faithfulness  

   Use the textbook’s actual definitions, structures, and interfaces. Do not redesign them.
   
2. Completeness  

   Include all declarations needed so the TARGET THEOREM parses and typechecks in a fresh project (Lean + Mathlib only).
   
3. Dependency-minimality  

   Include only declarations in the TRANSITIVE DEPENDENCY of the target theorem (statement + typechecking), not unrelated chapter material.
   
4. Compilation  

   The final snippet must be a single Lean file that compiles under the pinned Lean + Mathlib version.
   
---------------------

GIVEN / TOOLS

---------------------

You are always given:

- The full Lean source file for the chapter defining the target theorem.

- A textual “reference summary” of earlier declarations. This summary is DOCUMENTATION ONLY: you must never synthesize code from it.

You can call:

- `file\_look\_up\_tool` to fetch exact Lean source of other sections/chapters.

- `compile\_lean\_code\_tool` to test compilation of your current snippet.

All code you include for textbook symbols must ultimately be copied from real source files, never invented from the summary.

---------------------------------------

USAGE-BASED DEPENDENCY RULE

---------------------------------------

Let the “target theorem” be the one explicitly requested.

Define the USED NAME SET as:

- Every non-notation name in the target theorem’s statement.

- Every name appearing in the transitive dependency of declarations you already included, where that dependency is required for parsing, typechecking, instance/coercion / notation resolution, or unfolding to the level used by the textbook.

For ANY textbook name N in the USED NAME SET:

- You MUST obtain its real declaration from the project (current chapter file or via `file\_look\_up\_tool`).

- You MUST copy its full Lean declaration: name, binders, type, body, attributes, and any `sorry` in the original.

This includes core chapter objects like `Sequence`, `Sequence.ofNatFun`, `Sequence.from`, `Sequence.IsCauchy`, `CauchySequence`, `Real`, `LIM`, the order/abs structure on `Real`, and any instances/coercions they depend on. Do NOT simplify signatures (e.g. do NOT redefine `Sequence.Equiv` with a different domain, and do NOT replace quotient-based `Real` with `$\mathbb{Q}$`).

You must NOT import `Analysis.Section\_*`. Every textbook declaration you need must be physically copied into the snippet.

----------------

STRICT RULES

----------------

1. No fake textbook content

   - Never invent placeholder implementations for textbook names (e.g. `def Real := $\mathbb{Q}$`, `def foo := 0`, `abbrev bar := fun \_ => 0`, etc.).
   
   - Never change the type of a textbook symbol. Copied declarations must match source tokens (up to harmless whitespace).
   
2. Source of truth

   - For any textbook symbol you include, the body must come from the real Lean source, not from the textual summary and not from your own reconstruction.
   
   - You may reorder declarations slightly if needed for dependencies, but not modify their content.
   
3. Handling tactic-based proofs

   - If a textbook declaration comes with a proof term (including `by aesop`, `by grind`, etc.), you should:
   
     * Prefer copying the proof term EXACTLY as in source, AND copy any additional textbook lemmas/instances it depends on (also verbatim) when that does not explode the snippet.
     
     * If that proof relies on a large in-context lemma ecosystem that would balloon the context, you MAY replace ONLY THE PROOF TERM with `by sorry` **provided**:
     
       - The declaration’s type is a proposition (`theorem` / `lemma` / `example`), not a data definition.
       
       - The type (statement) is copied exactly from the source.
       
   - You MUST NOT replace the bodies of core data/structure/definition declarations (like the definition of `Real`, `Sequence`, `LIM`, etc.) with `sorry` or trivial placeholders.
   
4. Allowed and disallowed `sorry`

   - You must preserve any `sorry` that appears in the textbook source.
   
   - You MAY introduce NEW `sorry` **only** as proof terms of propositions (lemmas/theorems) where the original file gave a real proof and including its full dependency tree would be too large.
   
   - You must NOT introduce new `sorry` inside definitions/structures/instances that define data or operations.
   
5. New auxiliary declarations

   - You may add NEW auxiliaries only if:
   
     - Their names clearly indicate they are new (e.g. `Aux\_*`),

     - They do not shadow textbook names,
     
     - They are used solely to glue existing textbook declarations into a compilable file.
     
   - New auxiliaries must be fully defined without `sorry`.
   
6. Imports

   - You may import only Mathlib modules actually used (e.g. `Mathlib.Tactic`, `Mathlib.Data.Real.Basic`, etc.).
   
   - You must NOT import any textbook modules such as `Analysis.Section\_*`.
   
   - Assume no textbook code exists except what you copy into this single file.
   
7. No “I give up” prose

   - Your final answer MUST be a single Lean file inside one ```lean code block and NOTHING ELSE.
   
   - Do NOT output meta-text like “we could not assemble a self-contained slice”, “this exceeds the scope”, or any similar “I give up” header.
   
   - If tool limits prevent full resolution, still output the best faithful Lean snippet you can under the above rules.
   
----------------

WORKFLOW

----------------

1. Analyze the target theorem

   - List all non-notation names in its statement. Initialize the USED NAME SET with them.
   
   - Classify each as:
   
     (a) from the current chapter file,
     
     (b) from earlier textbook sections,
     
     (c) from Mathlib / core Lean.
     
2. Pull declarations from the current chapter

   - For each (a) in the USED NAME SET, locate its declaration and copy it verbatim.
   
   - Add any newly required names (from their types/bodies/instances/notation) into the USED NAME SET.
   
3. Pull declarations from other chapters

   - For each (b) in the USED NAME SET, call `file\_look\_up\_tool` for the relevant section and copy only the needed declarations verbatim.
   
   - Again, extend the USED NAME SET with whatever is needed to parse/typecheck those declarations.
   
4. Imports and namespaces

   - Add minimal Mathlib imports required for the copied code.
   
   - Recreate the namespace structure exactly as in the source (`namespace Chapter5`, `namespace Real`, etc.).
   
5. Compilation loop

   - Call `compile\_lean\_code\_tool` with the full snippet.
   
   - For unknown identifiers/namespaces: add the missing textbook declarations (via copying) or Mathlib imports.
   
   - For type/instance/coercion/notation errors: ensure you have copied the relevant instances/notations and that all declarations match the original types.
   
   - For tactic failures due to missing lemmas: either pull the missing lemmas verbatim, or (if that would be too large) change ONLY that lemma/theorem’s proof to `by sorry`, keeping the exact statement.
   
   Repeat until the snippet compiles or you hit tool limits; always maintain the rules above.
   
----------------

FINAL OUTPUT

----------------

- Output ONLY a single Lean file in a ```lean code block. No natural-language text before or after.

- The file must be self-contained (plus Mathlib imports) and, as far as your tools allow, compile successfully under the pinned Lean + Mathlib version.

}
\end{tcolorbox}
\twocolumn

\onecolumn
\subsubsection{Query Prompt for Textbook Extraction}
\label{box:tao-Query-Prompt}
\begin{tcolorbox}[
    colframe=black, 
    colback=white, 
    coltitle=white, 
    colbacktitle=black, 
    title=Query Prompt for Textbook Extraction,
    fonttitle=\bfseries,
    boxrule=1pt,
    arc=2mm,    
    left=4mm, right=4mm, top=3mm, bottom=3mm, 
    enhanced, 
    fontupper=\small,
    breakable
]

{
\setlength{\parskip}{5pt}
We need a **self-contained, compilable Lean snippet** for the following theorem

from {query['chapter\_name']}:

----- TARGET THEOREM -----

{query['query\_text']}

--------------------------

----- DEPENDENCY SET -----

{dependency\_set}

--------------------------

Your job is NOT to design new mathematics. Your job is to EXTRACT a faithful, dependency-minimal slice of the textbook’s Lean code so that the **target theorem** parses and typechecks **in isolation**, in a fresh Lean + Mathlib project, **WITHOUT** importing any `Analysis.Section\_*` modules.

Our goal is **context extraction**. Assume the target theorem is provable in the original project. The target theorem in the extracted snippet must **always** end with `by sorry` regardless of whether an external prover can solve it.

You must implement the following constraints:

============================================================

CORE RULES FOR BUILDING THE SELF-CONTAINED SNIPPET (V6)

============================================================

1. **Usage-based dependency rule**

   Build the USED NAME SET consisting of:
   
   - Every non-notation name in the target theorem's statement.
   
   - Every name appearing in the *transitive dependency* of included declarations required for parsing / typechecking / instance resolution / coercions / notation.
     
2. **For every textbook symbol in the USED NAME SET**, you must:

   - Obtain its **exact declaration** from the real source (the chapter file below, or any earlier section via `file\_look\_up\_tool`).
     
   - **Copy its Lean code verbatim**: type, body, attributes, proofs, and any `sorry` present in the textbook.
     
   - NEVER synthesize definitions from the summary below.
   
3. **The reference summary provided is documentation only.**

   - It may tell you *which* names exist, but *none* of its code may be used.
   
   - If you need a declaration, fetch its *actual* source file and copy it exactly.
   
4. **Handling textbook tactic proofs (e.g. by aesop, grind, simp, etc.)**

   - If feasible, copy the tactic proof term **exactly** and include any dependent
     textbook lemmas/instances (verbatim).
     
   - If including those dependencies excessively enlarges the snippet, you may replace the **proof term only** with `by sorry`, *but only for theorems/lemmas*.
     
   - You may NEVER `sorry` the bodies of definitions/structures/instances.
   
5. **Forbidden behaviors**

   - No placeholder implementations for textbook names (`:= 0`, `:= fun \_ => 0`, etc.).
   
   - No redefining textbook objects (e.g. replacing quotient-based `Real` with $\mathbb{Q}$).
   
   - No changing the signature of any textbook declaration.
   
   - No synthesized or “simplified” versions of textbook interfaces.
   
   - No “I give up” headers or meta explanations.
   
6. **Imports**

   - You may import relevant Mathlib modules (`Mathlib.Tactic`, `Mathlib.Data.Real.Basic`, …).
   
   - You must NOT import ANY textbook section (no `Analysis.Section\_*`).
   
7. **Target theorem**

   - Must retain EXACT name and statement.
   
   - Must end with `by sorry` **always** (even if a proof is known or found elsewhere).
   
8. **Final output**

   - A SINGLE Lean file, inside one ```lean code block.
   
   - No natural-language text before or after the code block.
   
   - The snippet must be self-contained and compilable under the pinned Mathlib/Lean version.
   
============================================================

AUTHORITATIVE SOURCE FOR THIS CHAPTER

============================================================

----- BEGIN SOURCE {query['chapter\_name']} -----

{source\_file}

----- END SOURCE {query['chapter\_name']} -----

============================================================

REFERENCE SUMMARY (FOR INFORMATION ONLY — DO NOT COPY CODE)

============================================================

{query['content']}

============================================================

Construct the final Lean snippet accordingly.
}
\end{tcolorbox}
\twocolumn

\onecolumn
\subsubsection{Initial Prompt for the Rewriting Stage of MathLib Translation}
\label{box:tao-Initial-Rewriting-Prompt}
\begin{tcolorbox}[
    colframe=black, 
    colback=white, 
    coltitle=white, 
    colbacktitle=black, 
    title=Initial Prompt for the Rewriting Stage of MathLib Translation,
    fonttitle=\bfseries,
    boxrule=1pt,
    arc=2mm,    
    left=4mm, right=4mm, top=3mm, bottom=3mm, 
    enhanced, 
    fontupper=\small,
    breakable
]

{
\setlength{\parskip}{5pt}

```lean4

\{TAOBENCH\_EXERCISE\}

```

The above is a Lean formalization exercise of a problem from Terence Tao's Analysis I. Many of the concepts in the given Lean code use local definition / local notation.

Please complete the following tasks:

1. First, understand and analyze the problem: read the given Lean code and the problem statement, and explain what the problem means mathematically.

2. Then, restate the problem using only standard definitions from Mathlib (no need to provide a proof). That is, rewrite the original problem, which uses local definitions / local notation, into a version that uses only the existing standard definitions in Mathlib. Only keep the formal statement of the problem (i.e. the Lean theorem / proposition statement), and do not give any proof.

In the Lean code, you may only use the following imports and settings:

import Mathlib

import Aesop

set\_option maxHeartbeats 0

open BigOperators Real Nat Topology Rat Set Filter

Do not add any other imports, and do not introduce any new local definitions or notation.

Please first give your analysis, and then provide the rewritten Lean problem statement (theorem declaration) using only the above restrictions, without any proof.

The theorem must keep the same name and end with := by sorry. You must ensure that this Lean code can be compiled in Lean4 and that the mathematical problem it describes is consistent with Terence Tao's exercise I gave you.

At the very end of your response, include your final Mathlib version of the problem in the following format (replace {Mathlib Version of the Problem} with your theorem declaration).

\#\#\# Mathlib Version:

```lean4

import Mathlib

import Aesop

set\_option maxHeartbeats 0

open BigOperators Real Nat Topology Rat Set Filter

namespace New\_Namespace

{Mathlib Version of the Problem}

end New\_Namespace

}
\end{tcolorbox}
\twocolumn

\onecolumn
\subsubsection{Prompt for Iterative Compilation Error Correction in the Rewriting Stage}
\label{box:tao-Error-Correction-Prompt}
\begin{tcolorbox}[
    colframe=black, 
    colback=white, 
    coltitle=white, 
    colbacktitle=black, 
    title=Prompt for Iterative Compilation Error Correction in the Rewriting Stage,
    fonttitle=\bfseries,
    boxrule=1pt,
    arc=2mm,    
    left=4mm, right=4mm, top=3mm, bottom=3mm, 
    enhanced, 
    fontupper=\small,
    breakable
]

{
\setlength{\parskip}{5pt}
```lean4

\{TAOBENCH\_EXERCISE\}

```

The above is a Lean formalization exercise of a problem from Terence Tao's Analysis I. Many of the concepts in the given Lean code use local definition / local notation.

Please complete the following tasks:

1. First, understand and analyze the problem: read the given Lean code and the problem statement, and explain what the problem means mathematically.

2. Then, restate the problem using only standard definitions from Mathlib (no need to provide a proof). That is, rewrite the original problem, which uses local definitions / local notation, into a version that uses only the existing standard definitions in Mathlib. Only keep the formal statement of the problem (i.e. the Lean theorem / proposition statement), and do not give any proof.

In the Lean code, you may only use the following imports and settings:

import Mathlib

import Aesop

set\_option maxHeartbeats 0

open BigOperators Real Nat Topology Rat Set Filter

Do not add any other imports, and do not introduce any new local definitions or notation.

Please first give your analysis, and then provide the rewritten Lean problem statement (theorem declaration) using only the above restrictions, without any proof.

The theorem must keep the same name and end with := by sorry. You must ensure that this Lean code can be compiled in Lean4 and that the mathematical problem it describes is consistent with Terence Tao's exercise I gave you.

\#\#\# This is your latest version of the answer:

```lean4

\{LATEST\_RESULT\}

```

\#\#\# However, it has compilation errors. The error message is:

\{ERROR\_MESSAGES\}

\#\#\# Please fix the errors while maintaining the mathematical meaning.

At the very end of your response, include your final Mathlib version of the problem in the following format (replace \{Mathlib Version of the Problem\} with your theorem declaration).

\#\#\# Mathlib Version:

```lean4

import Mathlib

import Aesop

set\_option maxHeartbeats 0

open BigOperators Real Nat Topology Rat Set Filter

namespace New\_Namespace

\{Mathlib Version of the Problem\}

end New\_Namespace

}
\end{tcolorbox}
\twocolumn

\onecolumn
\subsubsection{Prompt for the Equivalence Checking Stage of MathLib Translation}
\label{box:tao-Equivalence-Checking-Prompt}
\begin{tcolorbox}[
    colframe=black, 
    colback=white, 
    coltitle=white, 
    colbacktitle=black, 
    title=Prompt for the Equivalence Checking Stage of MathLib Translation,
    fonttitle=\bfseries,
    boxrule=1pt,
    arc=2mm,    
    left=4mm, right=4mm, top=3mm, bottom=3mm, 
    enhanced, 
    fontupper=\small,
    breakable
]

{
\setlength{\parskip}{5pt}

You are an expert in Lean4 and mathlib, and you are familiar with how the same mathematical statement can be represented in different formalizations, with different inductive definitions, notations, or contexts.

I will give you two versions of what is supposed to be the same exercise from Terence Tao's *Analysis I*:

1. Terence Tao exercise (Lean4 code)

2. Terence Tao goal state (the final statement to be proved)

3. Mathlib version exercise (Lean4 code)

4. Mathlib version goal state (the final statement to be proved)

Your task is to decide whether these two versions are proving the same mathematical statement.

\#\#\# Terence Tao exercise:

```lean4

\{TAOBENCH\_EXERCISE\}

```

\#\#\# Terence Tao goal state:

\{STATE\_OF\_TAOBENCH\_EXERCISE\}

\#\#\# Mathlib version exercise:

```lean4

\{TAOBENCHMATHLIB\_EXERCISE\}

```

\#\#\# Mathlib version goal state:

\{STATE\_TAOBENCHMATHLIB\_EXERCISE\}

Please proceed in the following steps:

1. First, understand and analyze the two versions separately: read the given Lean code of each exercise and its corresponding goal state to be proved, and explain what the exercise means mathematically.

2. Then, compare the two versions. Check whether the assumptions match logically (allowing for renaming of variables and standard equivalences), check whether the conclusions match logically, and then decide whether they are the same mathematical statement or if there is a real difference.

At the very end of your response, give your final answer of "Yes" or "No" in the following format (replace `\{Your final decision\}` with "Yes" if they are mathematically equivalent, or "No" if they are not mathematically equivalent).

\#\#\# Mathematical equivalence:

\{Your final decision\}

}
\end{tcolorbox}
\twocolumn

\end{document}